\documentclass[11pt,letterpaper]{article}
\usepackage{ijcnlp2017}
\usepackage{times}
\usepackage[utf8]{inputenc}
\usepackage[T1]{fontenc}
\usepackage{times}
\usepackage{helvet}
\usepackage{courier}
\usepackage{afterpage}
\usepackage{amsmath}
\usepackage{amsfonts}
\usepackage{amsthm}
\usepackage{latexsym}
\usepackage{graphicx}
\usepackage{xcolor}
\usepackage{url}
\usepackage{microtype}
\usepackage{listings}
\usepackage{paralist}
\ijcnlpfinalcopy

\newcommand{\ignore}[1]{}
\newcommand{\kw}[1]{\textcolor{red}{#1}}
\newcommand{\adam}[1]{\textcolor{orange}{}}
\newcommand{\ken}[1]{\textcolor{green}{}}

\title{Counterfactual Language Model Adaptation for Suggesting Phrases}

\author{Kenneth C. Arnold \\ Harvard CS\\Cambridge, MA
   \And Kai-Wei Chang \\ University of California\\Los Angeles, CA
   \And Adam T. Kalai \\ Microsoft Research\\Cambridge, MA}

\begin{document}

\maketitle

\begin{abstract}
Mobile devices use language models to suggest words and phrases for use in text entry. Traditional language models are based on contextual word frequency in a static corpus of text. However, certain types of phrases, when offered to writers as suggestions, may be systematically chosen more often than their frequency would predict. In this paper, we propose the task of generating suggestions that writers accept, a related but distinct task to making accurate predictions. Although this task is fundamentally interactive, we propose a counterfactual setting that permits offline training and evaluation. We find that even a simple language model can capture text characteristics that improve acceptability.

\end{abstract}

\section{Introduction}
\label{sec:intro}

\begin{figure}[t]
\centering\includegraphics[width=.9\columnwidth,trim={0 2cm 0 0.1cm},clip]{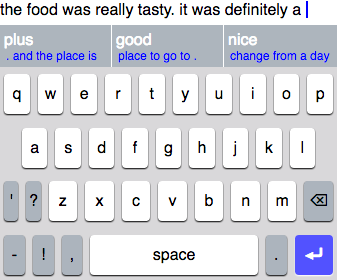}
\caption{\label{fig:ui}
We adapt a language model to offer
\emph{suggestions} during text composition.
In above example, even though the middle suggestion is predicted to be about 1,000 times more likely than the one on the right, a user prefers the
right one.
}
\end{figure}

Intelligent systems help us write by proactively suggesting words or phrases while we type.
These systems often build on a language model that picks \emph{most likely}
phrases based on previous words in context, in an attempt to increase entry speed and accuracy. However, recent work~\cite{megasuggestUI} has shown that writers appreciate suggestions that have creative wording, and can find phrases suggested based on frequency alone to be boring.
For example, at the beginning of a restaurant review, ``I love this place'' is a reasonable \emph{prediction}, but a review \emph{writer} might prefer a suggestion of a much less likely phrase such as ``This was truly a wonderful experience''---they may simply not have thought of this more enthusiastic phrase.
Figure~\ref{fig:ui} shows another example.

We propose a new task for NLP research: generate \emph{suggestions} for writers. Doing well at this task requires innovation in language generation but also interaction with people: suggestions must be evaluated by presenting them to actual writers. Since writing is a highly contextual creative process, traditional batch methods for training and evaluating human-facing systems are insufficient: asking someone whether they think something would make a good suggestion in a given context is very different from presenting them with a suggestion in a natural writing context and observing their response. But if evaluating every proposed parameter adjustment required interactive feedback from writers, research progress would be slow and limited to those with resources to run large-scale writing experiments.

In this paper we propose a hybrid approach: we maintain a natural human-centered objective, but introduce a proxy task that provides an unbiased estimate of expected performance on human evaluations. Our approach involves developing a \emph{stochastic} baseline system (which we call the \emph{reference policy}), logging data from how writers interact with it, then estimating the performance of candidate policies by comparing how they would behave with how the reference policy did behave in the contexts logged. As long as the behavior of the candidate policy is not too different from that of the reference policy (in a sense that we formalize), this approach replaces complex human-in-the-loop evaluation with a simple convex optimization problem.

This paper demonstrates our approach: we collected data of how humans use suggestions made by a reference policy while writing reviews of a well-known restaurant. We then used logged interaction data to optimize a simple discriminative language model, and find that even this simple model
generates better suggestions than a baseline trained without interaction data.
We also ran simulations to validate the estimation approach under a known model of human behavior.

Our contributions are summarized below:

\begin{compactitem}
	\item We present a new NLP task of {\em phrase suggestion} for writing.\footnote{Code and data are available at \url{https://github.com/kcarnold/counterfactual-lm}.}
	\item We show how to use counterfactual learning for goal-directed training of language models from interaction data.%
	\item We show that a simple discriminative language model can be trained with offline interaction data to generate better suggestions in unseen contexts.
\end{compactitem}

\section{Related Work}
Language models have a long history and  play an important role in many NLP applications~\cite{sordoni15conversation,rambow2001natural,mani-book01,JohnsonSLKWCTVW16}.
However, these models do not model human preferences from interactions.
Existing deployed keyboards use n-gram language models~\cite{Quinn2016:CostBenefitSuggestion,kneser1995improved}, or sometimes neural language models~\cite{kim2016character}, trained to predict the next word given recent context. Recent advances in language modeling have increased the accuracy of these predictions by using additional context~\cite{mikolov2012context}. But as argued in \citet{megasuggestUI}, these increases in accuracy do not necessarily translate into better suggestions.

The difference between suggestion and prediction is more pronounced when showing phrases rather than just words. Prior work has extended predictive language modeling to phrase prediction \cite{nandi2007effective} and sentence completion \cite{bickel2005learning}, but do not directly model human preferences. Google's ``Smart Reply'' email response suggestion system~\cite{smartreply2016} avoids showing a likely predicted response if it is too similar to one of the options already presented, but the approach is heuristic, based on a priori similarity. Search engine query completion also generates phrases that can function as suggestions, but is typically trained to predict what query is made (e.g., ~\citet{Jiang:2014:LUR:2600428.2609614}).

\section{Counterfactual Learning for Generating Suggestions}

We consider the task of generating good words and phrases to present to writers.
We choose a pragmatic quality measure: a suggestion system is good if \emph{it generates suggestions that writers accept}. Let $h$ denote a suggestion system, characterized by $h(y\vert{}x)$, the probability that $h$ will suggest the word or phrase $y$ when in context $x$ (e.g., words typed so far).\footnote{Our notation follows \citet{Swaminathan2015counterfactual} but uses ``reward'' rather than ``loss.'' Since $h(y\vert{}x)$ has the form of a contextual language model, we will refer to it as a ``model.''}
We consider deploying $h$ in an interactive interface such as Figure~\ref{fig:ui}, which suggests phrases using a familiar predictive typing interface. Let $\delta$ denote a reward that a system receives from that interaction; in our case, the number of words accepted.\footnote{Our setting admits alternative rewards, such as the speed that a sentence was written, or an annotator's rating of quality.} We define the overall quality of a suggestion system by its expected reward $E[\delta]$ over all contexts.

%

\iffalse
%
%
Unlike the domain or speaker adaptation setting, we do not assume
the existence of an existing (even small) set of annotations that label parts of existing writing that would make good suggestions for future writing. Instead, we seek to identify generalizable characteristics of words or phrases that make them desirable suggestions, and favor generating language with those characteristics through in situ interaction with humans. %
\fi

Counterfactual learning allows us to evaluate and ultimately learn models that differ from those that were deployed to collect the data, so we can deploy a single model and improve it based on the data collected~\cite{Swaminathan2015counterfactual}.
Intuitively, if we deploy a model $h_0$ and observe what actions it takes and what feedback it gets, we could improve the model by making it more likely to suggest the phrases that got good feedback.

 Suppose we deploy a \emph{reference model}\footnote{Some other literature calls $h_0$ a \emph{logging policy}.} $h_0$ and log a dataset
$$\mathcal{D}=\{(x_1,y_1,\delta_1,p_1),\ldots,(x_n,y_n,\delta_n,p_n)\}$$ of contexts (words typed so far), actions (phrases suggested), rewards, and propensities respectively, where $p_i\equiv{} h_0(y_i\vert{}x_i)$. Now consider deploying an alternative model $h_\theta$ (we will show an example as Eq. \eqref{eq:rewardp} below). We can obtain an unbiased estimate of the reward that $h_\theta$ would incur using importance sampling:
$$\hat{R}(h_\theta) = \frac{1}{n}\sum_{i=1}^{n}\delta_i {h_\theta(y_i\vert{}x_i)}/{p_i}.$$
However, the variance of this estimate can be unbounded because the importance weights $h_\theta(y_i\vert{}x_i) / p_i$ can be arbitrarily large for small $p_i$. Like~\citet{ionides2008truncated}, we clip the importance weights to a maximum $M$:
$$\hat{R}^M(h) = \frac{1}{n}\sum\nolimits_{i=1}^{n}\delta_i \min\left\{M, {h_\Theta(y_i\vert{}x_i)}/{p_i}\right\}.$$
The improved model can be learned by optimizing $$\hat{h_\theta}=\operatorname{argmax}_h \hat{R}^M(h).$$ This optimization problem is convex and differentiable; we solve it with BFGS.~\footnote{We use the BFGS implementation in SciPy. } %

\section{Demonstration Using Discriminative Language Modeling}

We now demonstrate how counterfactual learning can be used to evaluate and optimize the acceptability of suggestions made by a language model.
We start with a traditional predictive language model $h_0$ of any form, trained by maximum likelihood on a given corpus.\footnote{The model may take any form, but n-gram~\cite{KenLM-Heafield-estimate} and neural language models (e.g., \cite{kim2016character}) are common, and it may be unconditional or conditioned on some source features such as application, document, or topic context.}
This model can be used for generation: sampling from the model yields words or phrases that match the frequency statistics of the corpus.
However, rather than offering representative samples from $h_0$, most deployed systems instead sample from $p(w_i) \propto{} h_0(w_i)^{1/\tau}$, where $\tau$ is a ``temperature'' parameter;  $\tau=1$ corresponds to sampling based on $p_0$ (soft-max), while $\tau\rightarrow0$ corresponds to greedy maximum likelihood generation (hard-max), which many deployed keyboards use~\cite{Quinn2016:CostBenefitSuggestion}. The effect is to skew the sampling distribution towards more probable words.
This choice is based on a heuristic assumption that writers desire more probable suggestions; what if writers instead find common phrases to be overly cliché and favor more descriptive phrases?
To capture these potential effects, we add features that can emphasize various characteristics of the generated text, then use counterfactual learning to assign weights to those features that result in suggestions that writers prefer.

We consider locally-normalized log-linear language models of the form
\begin{equation}
\label{eq:rewardp}
h_\theta(y|x) = \prod_{i=1}^{|y|} \frac{\exp{\theta\cdot f(w_i|c, w_{[:i-1]})}}{\sum_{w'} \exp{\theta\cdot  f(w'|c, w_{[:i-1]})}},
\end{equation}
where $y$ is a phrase and $f(w_i|x, w_{[:i-1]})$ is a feature vector for a candidate word $w_i$ given its context $x$. ($w_{[:i-1]}$ is a shorthand for $\{w_1,w_2, \ldots w_{i-1}\}$.)
Models of this form are commonly used in sequence labeling tasks, where they are called Max-Entropy Markov Models~\cite{mccallum00memm}. Our approach generalizes to other models such as conditional random fields~\cite{lafferty01crf}.

The feature vector can include a variety of features. By changing feature weights, we obtain language models with different characteristics. To illustrate, we describe a model with three features below.
The first feature (\texttt{LM}) is the log likelihood under a base 5-gram language model $p_0(w_i|c, w_{[:i-1]})$ trained on the Yelp Dataset\footnote{\url{https://www.yelp.com/dataset_challenge}; we used only restaurant reviews} with Kneser-Ney smoothing~\cite{KenLM-Heafield-estimate}. The second and third features ``bonus'' two characteristics of $w_i$: \texttt{long-word} is a binary indicator of long word length (we arbitrarily choose $\ge{}6$ letters), and \texttt{POS} is a one-hot encoding of its most common POS tag. Table~\ref{tbl:example-gen} shows examples of phrases generated with different feature weights.

Note that if we set the weight vector to zero except for a weight of $1/\tau$ on \texttt{LM}, the model reduces to sampling from the base language model with ``temperature'' $\tau$. The fitted model weights of the log-linear model in our experiments is shown in supplementary material.

\begin{table}
{\small
\textbf{\texttt{LM} weight = 1, all other weights zero}:\\
 i didn't see a sign for; i am a huge sucker for

\textbf{\texttt{LM} weight = 1, \texttt{long-word} bonus = 1.0}:\\
another restaurant especially during sporting events

\textbf{\texttt{LM} weight = 1, \texttt{POS} adjective bonus = 3.0}:\\
great local bar and traditional southern%
}
\caption{\label{tbl:example-gen} Example phrases generated by the log-linear language model under various parameters. The context is the beginning-of-review token; all text is lowercased. Some phrases are not fully grammatical, but writers can accept a prefix.}

\end{table}

\paragraph{Reference model $h_0$.}
In counterfactual estimation, we deploy one reference model $h_0$ to learn another $\hat{h}$---but weight truncation will prevent $\hat{h}$ from deviating too far from $h_0$. So $h_0$ must offer a broad range of types of suggestions, but they must be of sufficiently quality that some are ultimately chosen. To balance these concerns, we use temperature sampling with a temperature $\tau=0.5$):
$$\frac{p_0(w_i|c, w_{[:i-1]})^{1/\tau}}{\sum_{w} p_0(w|c, w_{[:i-1]})^{1/\tau}}.$$ We use our reference model $h_0$ to generate 6-word suggestions one word at a time, so $p_i$ is the product of the conditional probabilities of each word.%

\begin{figure*}[t]
\centering
\includegraphics[width=\linewidth]{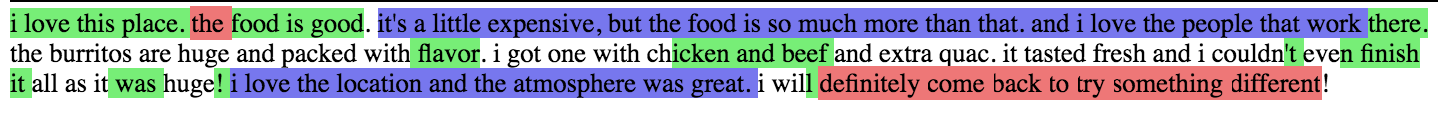}
\includegraphics[width=\linewidth]{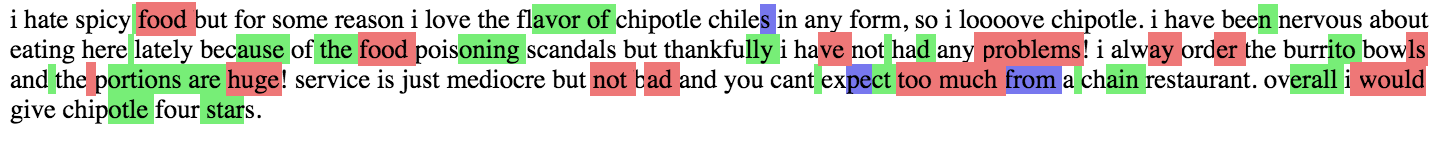}
\caption{\label{fig:example-review} Example reviews. A colored background indicates that the word was inserted by accepting a suggestion. Consecutive words with the same color were inserted as part of a phrase.}
\end{figure*}

\subsection{Simulation Experiment}
We present an illustrative model of suggestion acceptance behavior, and simulate acceptance behavior under that model to validate our methodology. Our method successfully learns a suggestion model fitting writer preference.

\paragraph{Desirability Model.}
We model the behavior of a writer using the interface in Fig. \ref{fig:ui}, which displays 3 suggestions at a time. At each timestep $i$ they can choose to accept one of the 3 suggestions $\{s_j^i\}_{j=1}^3$, or reject the suggestions by tapping a key. Let $\{p_j^i\}_{j=1}^3$ denote the likelihood of suggestion $s^i_j$ under a predictive model, and let $p^i_\emptyset=1-\sum_{j=1}^3 p^i_j$ denote the probability of any other word. Let $a^i_j$ denote the writer's probability of choosing the corresponding suggestion, and $a^i_j$ denote the probability of rejecting the suggestions offered. If the writer decided exactly what to write before interacting with the system and used suggestions for optimal efficiency, then $a_j^i$ would equal $p_j^i$. But suppose the writer finds certain suggestions \emph{desirable}. Let $D^i_j$ give the desirability of a suggestion, e.g., $D_j^i$ could be the number of long words in suggestion $s_j^i$. We model their behavior by adding the desirabilities to the log probabilities of each suggestion: $$a_j^{(i)} = p_j^{(i)}\exp({D_j^{(i)}})/Z^{(i)},\,a_\emptyset^{(i)}=p_\emptyset^{(i)}/Z^{(i)}$$  where $Z^{(i)}=1-\sum_j p_j^{(i)}(1-\exp(D_j^{(i)})).$
The net effect is to move probability mass from the ``reject'' action $a^i_\emptyset$ to suggestions that are close enough to what the writer wanted to say but desirable.

\begin{figure}[t]
\centering
\includegraphics[width=.9\columnwidth]{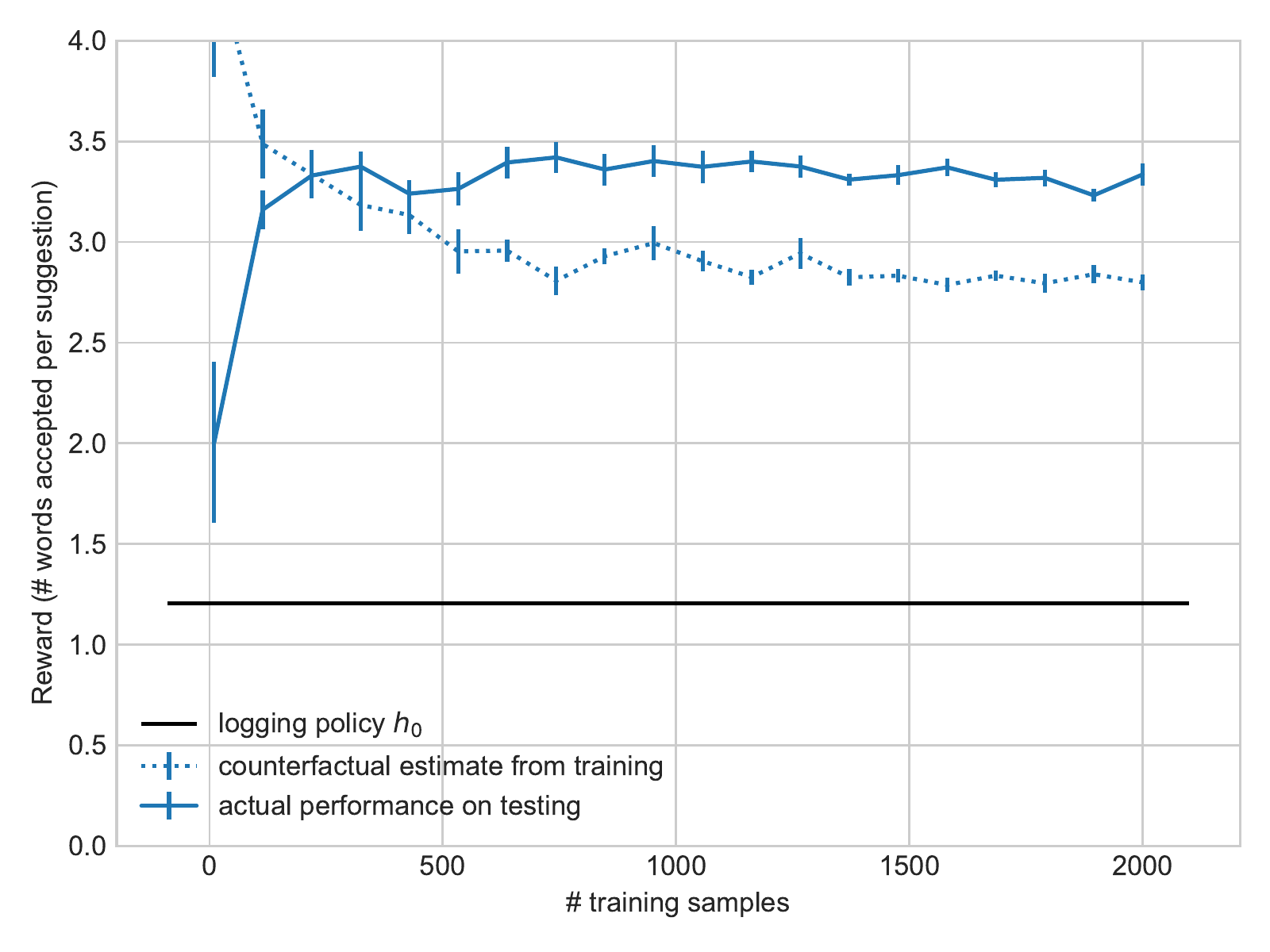}
\caption{\label{fig:sim-longword} We simulated learning a model based on the behavior of a writer who prefers long words, then presented suggestions from that learned model to the simulated writer. The model learned to make desirable predictions by optimizing the counterfactual estimated reward. Regularization causes that estimate to be conservative; the reward actually achieved by the model exceeded the estimate.}
\end{figure}

\paragraph{Experiment Settings and Results.}
We sample 10\% of the reviews in the Yelp Dataset, hold them out from training $h_0$, and split them into an equal-sized training set and test set. We randomly sample suggestion locations from the training set. We cut off that phrase and pretend to retype it. We generate three phrases from the reference model $h_0$, then allow the simulated author to pick one phrase, subject to their preference as modeled by the desirability model. We learn a customized language model and then evaluate it on an additional 500 sentences from the test set.

For an illustrative example, we set the desirability $D$ to the number of long words ($\ge 6$ characters) in the suggestion, multiplied by 10. Figure~\ref{fig:sim-longword} shows that counterfactual learning quickly finds model parameters that make suggestions that are more likely to be accepted, and the counterfactual estimates are not only useful for learning but also correlate well with the actual improvement.  In fact, since weight truncation (controlled by $M$) acts as regularization, the counterfactual estimate consistently \emph{underestimates} the actual reward.

\subsection{Experiments with Human Writers}
We recruited 74 workers through MTurk to write reviews of \emph{Chipotle Mexican Grill} using the interface in Fig \ref{fig:ui} from Arnold et al.~\shortcite{megasuggestUI}.
For the sake of simplicity, we assumed that all human writers have the same preference.
Based on pilot experiments, Chipotle was chosen as a restaurant that many crowd workers had dined at.
User feedback was largely positive, and users generally understood the suggestions' intent. The users' engagement with the suggestions varied greatly---some loved the suggestions and their entire review consisted of nearly only words entered with suggestions while others used very few suggestions. Several users reported that the suggestions helped them select words to write down an idea or also gave them ideas of what to write. We did not systematically enforce quality, but informally we find that most reviews written were grammatical and sensible, which indicates that participants evaluated suggestions before taking them.
The dataset contains 74 restaurant reviews typed with phrase suggestions. The mean word count is 69.3,  std=25.70. In total, this data comprises 5125 words, along with almost 30k suggestions made (including mid-word).

%

\iffalse
\begin{table}[t]
\centering
\begin{tabular}{@{}c|c@{ }c@{ }c@{ }c@{ }c@{ }c@{ }c@{}}
\# accepted&0&1&2&3&4&5&6\\
count&27,859&1,397&306&130&91&68&107\\
\end{tabular}
\caption{\label{table:num-accepted} Our dataset includes 29,958 suggestions made by the system during typing.
Authors accepted at least one word of 2099 suggestions (7\%), and at least 2
words in 702 suggestions (2.3\%). In total, 3745 out of 5125 words in the
corpus were entered using suggestions. These acceptance rates are comparable
with those observed in other work.}
%
\end{table}
\fi

\begin{figure}
\centering
\includegraphics[trim={0 0.5cm 0 1.5cm},clip,width=0.95\columnwidth]{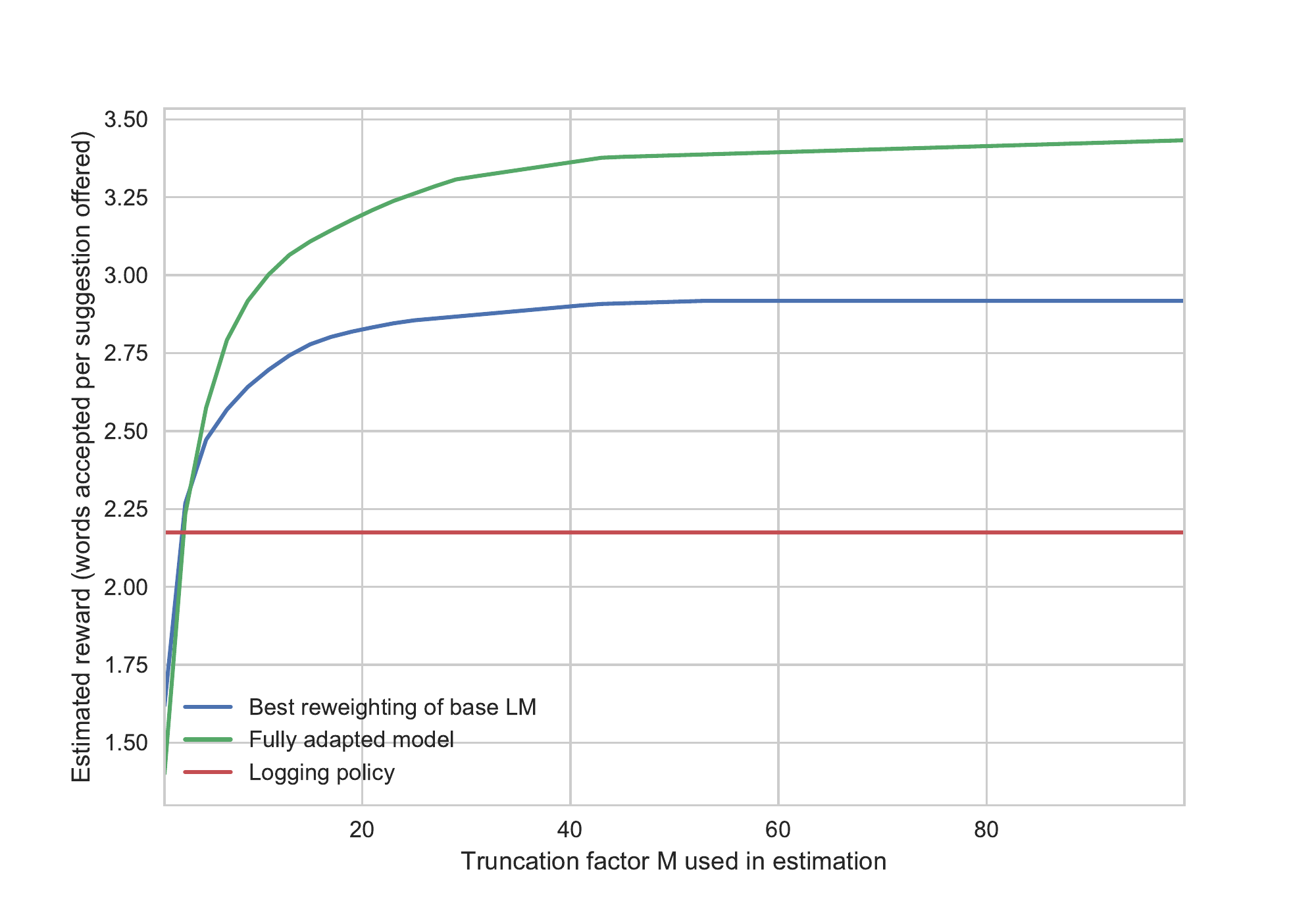}
\caption{\label{fig:est-reward-dataset} The customized model consistently improves expected reward over baselines (reference LM, and the best ``temperature'' reweighting LM) in held-out data. Although the result is an estimated using weight truncation at $M$, the improvement holds for all reasonable $M$. }
\end{figure}

\paragraph{Estimated Generation Performance.}
We learn an improved suggestion model by the estimated expected reward ($\hat{R}^M$). We fix $M=10$ and evaluate the performance of the learned parameters on held-out data using 5-fold cross-validation. Figure~\ref{fig:est-reward-dataset} shows that while the estimated performance of the new model does vary with the $M$ used when estimating the expected reward, the relationships are consistent: the fitted model consistently receives the highest expected reward, followed by an ablated model that can only adjust the temperature parameter $\tau$, and both outperform the reference model (with $\tau=1$). The fitted model weights \iffalse{}given in Supplemental Material\fi{} suggest that the workers seemed to prefer long words and pronouns, and eschewed punctuation.

\iffalse
We make several simplifying assumptions and leave these directions to future work. First, we assume that the distribution over contexts is fixed and not influenced by prior suggestions. How past suggestions will influence future writing is a fascinating issue
%
--- once a suggestion interface begins to dramatically influence the distribution over content generated, it has already achieved our motivating goal of having suggestions accepted. Second, practical interfaces such as the one we use in our experiments may offer more than one suggestion at a time; the formulation above assumes that suggestions do not interact.
%
Some related issues such as diversity of suggestions and building personalized system have been studied in information retrieval and machine learning (e.g., \cite{shokouhi2013learning,guzman2014efficiently}). \kw{Move this paragraph to Discussion?}

%
\fi

\section{Discussion}

Our model assumed all writers have the same preferences. Modeling variations between writers, such as in style or vocabulary, could improve performance, as has been done in other domains~(e.g., \citet{Lee:2017:PRL:3105672.3105673}). Each review in our dataset was written by a different writer, so our dataset could be used to evaluate online personalization approaches.

Our task of crowdsourced reviews of a single restaurant may not be representative of other tasks or populations of users. However, the predictive language model is a replaceable component, and a stronger model that incorporates more context (e.g., \citet{sordoni15conversation}) could improve our baselines and extend our approach to other domains. %

Future work can improve on the simple discriminative language model presented here to increase grammaticality and relevance, and thus acceptability, of the suggestions that the customized language models generate.

\paragraph{Acknowledgements}
Kai-Wei Chang was supported in part by National Science Foundation Grant IIS-1657193. Part of the work was done while Kai-Wei Chang and Kenneth C. Arnold visited Microsoft Research, Cambridge.

\bibliography{bibfile}
\bibliographystyle{ijcnlp2017}
\onecolumn{
\newpage

\appendix
\newpage
\onecolumn{
\section{Supplemental Material}
\subsection{Experiment details}
Crowd workers were U.S.-based Mechanical Turk workers who were paid \$3.50 to write a review of Chipotle using the keyboard interface illustrated in Figure \ref{fig:ui}. They could elect to use the interface on either a smartphone or on a personal computer. In the former case, the interaction was natural as it mimicked a standard keyboard. In the latter case, users clicked with their mouses on the screen to simulate taps. (There did not seem to be significant differences between these two groups.) The instructions are given below:

\begin{lstlisting}
Go to <URL> on your computer or phone.
Try out our new keyboard by pretending you're writing a restaurant review. For this part we just want you to play around -- it doesn't matter what you type as long as you understand how it works. Click the submit button and enter the code here: ____________

Did you use your phone or did you use you computer?
How would you describe the new keyboard to a friend? How do you use the phrase suggestions?

Next, please go back to <URL> on your computer or phone (reload if necessary).

Now please use the keyboard to write a fun review for Chipotle, the infamous chain Mexican restaurant. The ideal review is well written (entertaining, colorful, interesting), and has specific details about Chipotle menu items, service, atmosphere, etc. Please do not randomly click on nonsense suggestions -- we all know Chipotle doesn't serve pizza or burgers. We will bonus our favorite review!
\end{lstlisting}

\subsection{Qualitative feedback}
We include quotes from feedback from participants who used the system with suggestions generated by the reference model $h_0$, i.e., $n$-grams with temperature 1/2. Some users found the suggestions accurate at predicting what they intended to say, some found them useful in shaping one's thoughts or finding the ``right words,'' and others found them overly simplistic or irrelevant.

{\footnotesize
\begin{itemize}

\item ``The phrases were helpful in giving ideas of where I wanted to go next in my writing, more as a jumping off point than word for word. I've always liked predictive text so the phrases are the next level of what I never knew I wanted.''
\item ``Kind of easy to review but they also sometimes went totally tangent directions to the thoughts that I was trying to accomplish.''

\item ``I was surprised how well the words matched up with what I was expecting to type.''
\item ``I did like the phrase suggestions very much. They really came in handy when you knew what you wanted to say, but just couldn't find the right words.''

\item ``I thought they were very easy to use and helped me shape my thoughts as well!  I think they may have been a bit too simple in their own, but became more creative with my input.''

\end{itemize}
}

\subsection{Fitted model weights}

The following table gives the fitted weights for each feature in the log-linear model, averaged across dataset folds.

{\footnotesize

\begin{tabular}{lrrrrrrrrrrrrr}
{} &  base LM &  is\_long &  PUNCT &  ADJ &  ADP &  ADV &  CONJ &  DET &  NOUN &  NUM &  PRON &  PRT &  VERB \\\hline
mean &            2.04 &     0.92 &  -1.16 & 1.03 & 1.45 & 0.45 &  0.91 & 0.36 &  0.96 & 0.87 &  1.68 & 0.23 &  0.79 \\
std  &            0.16 &     0.14 &   0.26 & 0.61 & 0.38 & 0.55 &  0.26 & 0.22 &  0.14 & 0.27 &  0.20 & 1.00 &  0.32 \\
\end{tabular}
}

}
\end{document}